\documentclass[conference]{IEEEtran}
\IEEEoverridecommandlockouts
\usepackage{cite}
\usepackage{booktabs} 
\usepackage{amsmath,amssymb,amsfonts}
\usepackage{algorithmic}
\usepackage{float}
\usepackage{graphicx}
\usepackage{tabularx}
\usepackage{textcomp}
\usepackage{xcolor}
\usepackage[pdfborder={0 0 0}]{hyperref}
\usepackage{etoolbox}
\usepackage{tablefootnote}
\makeatletter
\patchcmd{\@makecaption}
  {\scshape}
  {}
  {}
  {}
\makeatother
\usepackage{caption}
\newcommand{\doweaccept}{} %

\def\BibTeX{{\rm B\kern-.05em{\sc i\kern-.025em b}\kern-.08em
    T\kern-.1667em\lower.7ex\hbox{E}\kern-.125emX}}
\begin{document}

\title{\textsc{OpenRTLSet}: A Fully Open-Source Dataset for Large Language Model-based Verilog Module Design\\
}
\author{\IEEEauthorblockN{Jinghua Wang$^*$, Lily Jiaxin Wan$^*$, Sanjana Pingali$^*$, Scott Smith$^*$, Manvi Jha$^*$, \\ Shalini Sivakumar, Xing Zhao, Kaiwen Cao, Deming Chen}
\IEEEauthorblockA{\textit{University of Illinois Urbana-Champaign} \\ \{jinghua3, wan25, pingali4, scottcs2, manvij2, ss251, xingz6, kaiwenc2, dchen\}@illinois.edu}
}

\maketitle
\def\thefootnote{*}\footnotetext{These authors contributed equally to this work.}\def\thefootnote{\arabic{footnote}}

\begin{abstract}


\textsc{OpenRTLSet}\footnote{Code available at {\color{blue} \url{https://github.com/UIUC-ChenLab/OpenRTLSet}}; dataset and fine-tined LLMs available at {\color{blue}\url{https://huggingface.co/datasets/ESCAD/OpenRTLSet}}.}
introduces the largest fully open-source dataset for hardware design, offering over 131,000 diverse Verilog code samples to the research community and industry. Our dataset uniquely combines Verilog code from GitHub repositories (102k modules), VHDL translations (5k modules), and synthesizable C/C++ translations (24k modules), all freely accessible without proprietary restrictions. Using the reasoning model DeepSeek-R1, we generated paired natural language descriptions for each code sample, enabling fine-tuning of various language model families (e.g., Qwen and Granite) for Verilog code generation. Our dataset explores multiple options, including Verilator-generated C++ files as additional context during labeling, quantization techniques (INT4 vs. BF16), and performance differences across model sizes (7B-32B parameters). 
\textsc{OpenRTLSet} demonstrates that open-source approaches can achieve superior performance in hardware design tasks, establishing a new foundation for accessible research and commercial use in this domain.

\end{abstract}


\section{Introduction}
Recent advancements in Large Language Models (LLMs), exemplified by systems such as CodeLlama~\cite{roziere2023code}, DeepSeek-Coder~\cite{guo2024deepseek} and Qwen-Coder~\cite{hui2024qwen2}, have redefined the landscape of automated code generation. These state-of-the-art models leverage extensive pretraining on massive, publicly available code corpora to internalize complex algorithmic patterns and API semantics, thereby enabling the generation of code in languages like Python~\cite{zhang2024pybench}\cite{pan2023stelocoder}, JavaScript~\cite{vadoce2024enhancing}\cite{le2024study}, and C++~\cite{yu2024large}.

Building on these successes, there is growing interest in extending LLM capabilities to hardware design—specifically, to the generation of Hardware Description Language (HDL) code from high-level specifications~\cite{wan2024software}\cite{huang2024new}\cite{xu2024llm}. However, applying LLMs to Verilog—the predominant HDL in digital circuit design—faces a critical challenge: the severely limited availability of large-scale, high-quality training datasets~\cite{pan2025survey}\cite{alsaqer2024potential}. Unlike software languages, which benefit from abundant open-source repositories, Verilog code is often restricted within specialized industrial domains. This scarcity, compounded by the proprietary nature of industrial hardware IP, has stifled progress in developing domain-specific models capable of generating syntactically and functionally correct Verilog code. The lack of accessible, standardized datasets remains a fundamental barrier to advancing LLM-driven automation in hardware engineering.

Efforts to bridge this gap have produced domain-specific LLMs like DAVE~\cite{pearce2020dave}, VeriGen~\cite{thakur2024verigen}, and CodeV~\cite{zhao2024codev}, which fine-tune general-purpose models on limited HDL datasets. While datasets such as VGen \cite{thakur2023benchmarking} aggregate code from GitHub and textbooks, evaluations reveal persistent failures in generating basic constructs like Linear-Feedback Shift Registers (LFSRs) and syntax-compliant modules \cite{blocklove2024evaluating}. Commercial tools like Synopsys.ai \cite{narayanan2023redefining} and ChipNeMo \cite{liu2024domain} address industrial needs but lack transparency. Several large-scale datasets like PyraNet~\cite{nadimi2024pyranet} and MG-Verilog~\cite{zhang2024mgverilog}, and proprietary collections from industry leaders such as NVIDIA and Synopsys have emerged. Still, they do not guarantee fully open-source licensing and, thus, are not viable for free commercial use. This creates significant barriers for researchers and companies looking to build competitive commercial LLM solutions. These limitations underscore the need for fully open-source, large-scale datasets paired with reproducible methodologies to advance LLM-driven hardware design that can be freely used in both academic and commercial settings.

In response to these challenges, we introduce \textsc{OpenRTLSet}.
We summarize our contributions as follows:
\begin{itemize}
    \item We introduce \textsc{OpenRTLSet}—\textbf{the largest fully open-source Verilog dataset} containing 131k diverse Verilog modules from GitHub-sourced designs, VHDL/SystemVerilog translations, and HLS-generated RTL. Unlike other large-scale open-source Verilog datasets that are not completely open for commercial use, \textsc{OpenRTLSet} ensures complete reproducibility through traceable GitHub origins and permissive licensing, directly addressing a critical barrier to \textbf{transparent research} and \textbf{commercial application} in LLM-driven hardware design.
        \item We develop a comprehensive \textbf{fully open-source end-to-end flow} for high-quality data labeling that integrates reasoning-driven and prompt-engineering techniques, optionally including Verilator-generated C++ pairing with the Verilog modules during natural language labeling. 
        By leveraging the DeepSeek-R1 model, we generate natural language descriptions of each Verilog module to form a high-quality dataset that can fine-tune various Coding LLMs into Verilog experts. 
        \item We conduct systematic evaluations of LLM performance on Verilog code generation. Through our experiments, we explore the effects of model quantization, dataset augmentation with cross-language Verilog representations, and structured prompt engineering techniques. Our comparative evaluation against existing datasets establishes \textsc{OpenRTLSet} as a definitive, high-quality benchmark for LLM-driven RTL generation research and application.
        \item Using the MG-Verilog original dataset with multi-granularity labeling as our baseline, we apply our labeling techniques to the complete 131k \textsc{OpenRTLSet} and fine-tune Qwen2.5-32B. This yields an absolute improvement of 5.7\% in Pass@1, 8.3\% in Pass@5 and 7.9\% in Pass@10 on the VerilogEval-Machine benchmark. On the VerilogEval-Human benchmark, we observe absolute gains of 9.7\% in Pass@1, 13.0\% in Pass@5, and 14.2\% in Pass@10. Across all evaluated LLMs from 7B to 32B and all Pass@k metrics, \textsc{OpenRTLSet} consistently outperforms MG-Verilog.

\end{itemize}

In the rest of the paper, Section~\ref{sec:background} reviews some relevant background and discusses related work in LLM-based hardware design and existing datasets. Section~\ref{sec:methodology} details our flow for creating \textsc{OpenRTLSet}, including data collection, parsing, and labeling. Section~\ref{sec:experiments} presents our experimental results and ablation studies comparing different model architectures fine-tuned on our dataset. Finally, Section~\ref{section:conclusion} concludes the study.

\section{Background \& Related Work}
\label{sec:background}

\subsection{Hardware Description Languages and Tools}
Numerous tools and languages can be used to express hardware designs. The focus of this dataset is the Verilog language.\footnote{Throughout this work, we state our target language as Verilog. However, our dataset is comprised of both Verilog and SystemVerilog modules.} Another popular hardware language is VHDL. Both languages can express the low-level details of hardware design, such as the gates, registers, and timing necessary for a correct and high-performance design. High Level Synthesis (HLS) is a popular technique for designing hardware, especially for FPGAs \cite{HLSforFPGAs, HLS-TRETS, hpca2022scalehls, dac2022scalehls, ye2023hida}. HLS involves describing a hardware design in a higher-level language such as C or C++ and then transforming that code into actual hardware (e.g., by generating Verilog). Our dataset leverages all three of the aforementioned methods for describing hardware. We convert non-Verilog code into Verilog by leveraging techniques discussed in Section \ref{sec:methodology}. 

Other hardware languages and tools, such as Chisel \cite{Chisel}, Veryl \cite{hatta2024verylnewhardwaredescription}, and Spade \cite{skarman2023spadeexpressionbasedhdlpipelines} exist but are significantly less popular. There were on the order of a few hundred GitHub repositories for these languages/tools compared to the tens of thousands of Verilog and VHDL repositories. Consequently, we ignored them during the creation of our dataset. We focused on the languages (Verilog, VHDL, and synthesizable C/C++) that would provide the most code samples. 

LLMs lack the extensive Verilog knowledge compared with programming languages such as Python, and C, C++. LLMs could generate better natural language descriptions of Verilog modules when provided with code from these programming languages as compared to Verilog. BetterV \cite{pei2024betterv} introduces the idea of converting the Verilog code into C code to help the LLM better understand the Verilog design provided. Our research advances this concept through an innovative application of Verilator, enabling specialized Verilog-to-C++ conversion that includes not only the Verilog but also the cross compiled C++ model. We analyze the effectiveness of labeling Verilog code with Verilator-generated C++ in Section \ref{sec:experiments}. 

\subsection{Hardware Design Datasets for LLMs} 
Recent advances in LLM-assisted hardware design have underscored the need for datasets that not only cover a wide spectrum of design examples but also provide multi-grained information—from high-level design intents to detailed block-level labels. For instance, MG-Verilog \cite{zhang2024mgverilog} demonstrates that enriching training data with diverse, granular design examples significantly improves the quality of generated Verilog code. 
PyraNet \cite{nadimi2024pyranet} introduces a multi-tiered approach to Verilog design abstraction, enhancing LLMs' ability to capture intricate design patterns across abstraction levels. However, PyraNet’s restrictive license limits its use in research and commercial applications. Commercial use would conflict with the ChatGPT terms of service \cite{OpenAITOS} (ChatGPT output should not be used to train/fine-tune competitive LLM products).
In contrast to existing approaches that either rely primarily on limited and simplistic RTL repositories or non-commercial-permissive source Verilog modules and/or flow, our work introduces a large-scale, fully open-source, and systematically curated collection of Verilog modules together with the dataset generation flow. 
Our dataset and flow are tailored for modular hardware design applications,  bridging the gap between academic benchmarks and practical, real-world challenges.


\section{Methodology}
\label{sec:methodology}

\subsection{GitHub Scraping}

\subsubsection{Data Collection}
\label{subsec:datacollection}

We leveraged a systematic and rigorous methodology to construct a high-quality dataset for fine-tuning a LLM capable of understanding and generating Verilog code. We initiated the process by conducting a comprehensive search on GitHub for open-source repositories containing Verilog, VHDL, and C/C++ files.
All discovered files were downloaded for further examination. 

We only kept code licensed as completely free for use in both \textbf{academic} and \textbf{commercial} settings. During our parsing, we found approximately 81k open-source repositories containing Verilog code, 5k open-source repositories containing VHDL code, and 103k open-source repositories containing C/C++ code. Other repositories did not have permissive licenses and were pruned from the dataset. 

\subsubsection{Language Translation}
In addition to directly using Verilog code, we also transformed VHDL and synthesizable C/C++ code into Verilog. Approximately 5k VHDL files were converted into Verilog. VHDL files were converted using an open-source repository called vhd2vl \cite{ldoolitt_vhd2vl}. VHDL files that failed conversion were not included in our dataset. 

C/C++ code was also collected from open source repositories on GitHub.  However, many C/C++ repositories contain non-synthesizable code that cannot be directly converted to hardware. So, we passed all collected C/C++ code through a custom regular expression checker. We pruned away code that cannot be synthesized due to dynamic memory allocation, pointers, while loops, recursion, and branch control loops such as jmp, goto, etc. After this additional filtering, around 12k repositories were left. We used these to obtain Verilog code by converting the C/C++ code using Vitis HLS. From these 12k only around 1k successfully generated Verilog modules through Vitis HLS. These failures are due to issues like certain C++ standard library calls that are not synthesizable (such as std::map). We obtained approximately 24k total Verilog modules from the synthesizable C/C++ code.


\subsubsection{Module Extraction}

Verilog files containing multiple modules were decomposed into individual modules to enhance learning granularity and reusability. This was achieved by parsing files for \textit{module} and \textit{endmodule} keywords and saving each module as a separate .v file. Single-module files help the LLM focus on the structure and syntax of individual components and support targeted fine-tuning or testing.

\subsubsection{Preprocessing}

To prepare the modules for fine-tuning, we applied several preprocessing steps aimed at cleaning, standardizing, and formatting the data. (1) We removed redundant elements such as comments and whitespace, (2) we ensured all code had uniform indentation and spacing, and (3) module names and I/O pins were extracted from the modules. 

\subsubsection{Data Structuring}

The preprocessed Verilog modules were organized into a structured dataset in the JSONL (JSON Lines) format, where each line represents a single Verilog module as a JSON object. This dataset is structured into two main fields: header and code. The `header' field contains key metadata about the module, including its name, derived from the module declaration, and its I/O pins, which summarize the input, output, and bidirectional signals associated with the module. The `code' field contains the cleaned and formatted Verilog, stripped of comments and unnecessary whitespace to focus solely on functional elements. This structured representation ensures compatibility with the LLM fine-tuning flow, enabling efficient processing and learning. Moreover, by separating metadata from functional code, the dataset enhances interpretability and supports diverse downstream applications.

\subsubsection{Data Cleaning}
\label{subsection:github-datacleaning}
Before testing with the LLM, the dataset undergoes a comprehensive cleaning process to ensure quality and relevance. Summaries that, when combined with custom sentence initiators, terminators, and code headers, may exceed the token length limit are removed, along with those containing repetitive or incomplete text. Repetition is detected using token matching, while incomplete summaries are identified through the absence of sentence-ending punctuation (e.g., `.') and addressed through manual review. Skeleton codes—modules containing only `module' and `endmodule' declarations without any functional logic—are also excluded. To further enhance consistency, all remaining data is uniformly formatted, ensuring compatibility with the testing flow and maximizing the dataset's effectiveness.

\subsection{Dataset Statistics}
Table \ref{tab:dataset_stats} provides a summary of the statistics for our dataset. The data is separated by input type (Verilog, VHDL, C/C++). 

\begin{table}[h]
    \centering
    \caption{Data collection statistics}
    \begin{tabular}{|c|c|c|c|} 
        \hline
        Language & Total Open-Source Repositories & Final Verilog Modules \\ 
        \hline
        Verilog & 81k & 102k \\
        \hline
        VHDL & 5k & 5k \\
        \hline
        C/C++ & 103k & 24k \\
        \hline %
    \end{tabular}
    \label{tab:dataset_stats}
\end{table}

\subsection{End-to-End Flow}
\begin{figure*}[t]
\centering
\includegraphics[width=1\linewidth]{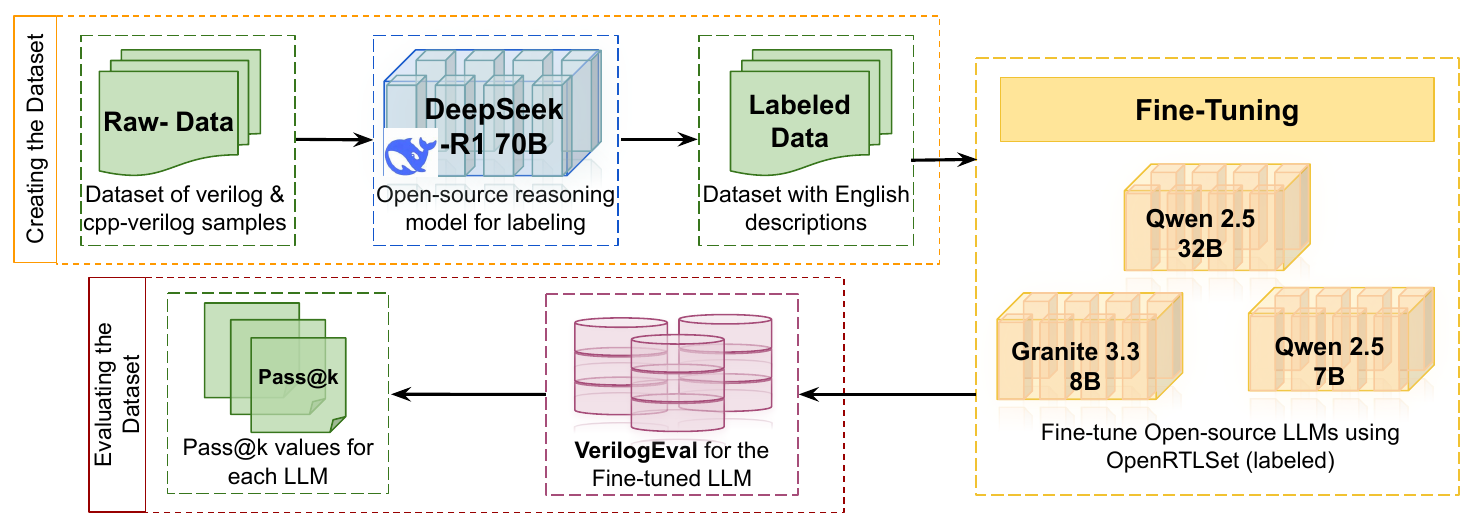}
\caption{The end-to-end flow for \textsc{OpenRTLSet} creation, labeling, fine-tuning, and evaluation. The process begins with raw data collection from GitHub repositories, followed by labeling using DeepSeek-R1 70B, fine-tuning of multiple LLM architectures, and evaluation using the VerilogEval benchmark.}
\label{fig:pipeline}
\end{figure*}

To effectively create, label, and evaluate our \textsc{OpenRTLSet} dataset, we developed a comprehensive end-to-end flow that spans from initial data collection to model evaluation. Figure \ref{fig:pipeline} illustrates this complete workflow. All labels were generated using tools that are completely open source, meaning the Verilog code and labels can be used in \textbf{academic} and \textbf{commercial} settings.  Many LLMs such as ChatGPT have explicit policies against using their output data to train competitor LLMs \cite{OpenAITOS}. We ensured that all utilized tools do not have such policies.

Our flow consists of three primary stages:
\subsubsection{Dataset Creation}
The process begins with raw data collection, as discussed in Section~\ref{subsec:datacollection}. 
\subsubsection{Dataset Labeling}
We leverage the DeepSeek-R1 70B model, an open-source reasoning-focused LLM, to generate high-quality natural language descriptions for each Verilog module. This labeling process incorporates our novel approach of using Verilator~\cite{snyder_veripool}-generated C++ context to enhance semantic understanding, resulting in more accurate and comprehensive descriptions. The labeled dataset pairs each module with its corresponding natural language description, creating training data suitable for fine-tuning various LLMs.
\subsubsection{Model Fine-Tuning and Evaluation}
Using the labeled dataset, we fine-tune multiple open-source language models, including the Granite series (8B) and Qwen series (7B and 32B). These fine-tuned models are then evaluated using the VerilogEval benchmark to assess their Verilog generation capabilities through Pass@k metrics. This standardized evaluation allows for fair comparison across different model architectures and training configurations.

This flow ensures complete reproducibility through traceable GitHub origins and permissive licensing. By maintaining transparency at each stage, we address a critical barrier in LLM-driven hardware design research. The systematic structure also facilitates ablation studies and comparative analyses, as detailed in the following section, allowing us to isolate the effects of different factors on model performance.
\section{Experiments}
\label{sec:experiments}

In this section, we evaluate the effectiveness of our \textsc{OpenRTLSet} dataset and benchmark various language models for Verilog code generation. Our experimental framework addresses two main objectives: (1) optimizing prompt engineering strategies for generating natural language descriptions of Verilog modules, and (2) comparing the performance of models fine-tuned on our dataset with existing benchmarks.

\subsection{Dataset Labeling Ablation Study} \label{4a}
To improve the quality of the labels used for fine-tuning, we conduct an ablation study focusing on three factors:

\subsubsection{Code Source Configuration}
We compare two settings: one where Verilator~\cite{snyder_veripool} generates both C++ and Verilog code to provide richer contextual cues, and another where only the original Verilog code is used. This experiment determines if the additional context from Verilator-generated C++ code enhances label quality.

\subsubsection{Code Enhancement Strategy} 
We evaluate whether enhancing Verilog modules with their Verilator C++ code improves downstream performance. Out of the 11k Verilog modules in MG-Verilog dataset, approximately 40\% of modules successfully pass the Verilator flow. In our ablation study, we downsample MG-Verilog to 10\% of its size (1k Verilog modules) using random selections. Among these 1k Verilog modules, around 40\% successfully pass the Verilator flow. We compare these labeling approaches: (1) augmenting Verilog modules with their corresponding C++ code generated via Verilator where available, (2) using only original Verilog code for all modules, and (3) using Verilog \& Verilator C++ for only around 400 samples that successfully pass Verilator flow. Our abalation study helps determine whether the additional semantic context provided by Verilator C++ code improves the model's understanding of hardware functionality and ultimately leads to better code generation capabilities. The conclusion of our ablation study guides our Comparative Benchmark Analysis in Section \ref{4b}.

\subsubsection{Quantization Settings}

For all labeling experiments, we use the reasoning-based DeepSeek-R1 70B model~\cite{guo2025deepseek} to generate natural language descriptions, which are then used to fine-tune our Verilog code generation models.
We compare the effects of INT4 and BF16 quantization on DeepSeek-R1 70B during labeling. This study aims to balance computational efficiency and labeling fidelity optimally.

\subsection{Comparative Benchmark Analysis} \label{4b}

We benchmark \textsc{OpenRTLSet} against the MG-Verilog dataset~\cite{zhang2024mgverilog}. Since MG-Verilog has 11,144 Verilog modules, we downsample our \textsc{OpenRTLSet} to 11,144 Verilog modules to fairly compare our dataset with MG-Verilog. Our analysis considers three configurations: 

\begin{enumerate} 
\item MG-Verilog with its original LLaMA-series labels. 
\item MG-Verilog relabeled using our DeepSeek-R1 70B labeling flow.
\item Downsampled \textsc{OpenRTLSet} labeled using our DeepSeek-R1 70B flow.
\end{enumerate} 

Comparing 1) and 2) enables evaluations on the effectiveness of the flow in MG-Verilog vs. our labeling flow; comparing 2) and 3) enables evaluations on the quality of the Verilog code in MG-Verilog vs. \textsc{OpenRTLSet}.

\subsection{Evaluation Protocol} \label{4c}
\subsubsection{Metrics}
Model performance is evaluated using Pass@1, Pass@5, and Pass@10 metrics on the VerilogEval benchmark~\cite{liu2023verilogevalevaluatinglargelanguage}. These metrics evaluate syntax compliance, functional correctness, and overall code quality at a generation temperature of 0.7.

\subsubsection{Model and Hardware Configuration}
We fine-tune two families of models—Granite-Code-Instruct (8B)\cite{granite2024granite} and Qwen2.5-Coder (7B and 32B parameters)\cite{yang2024qwen2}—with hyperparameters such as batch size, learning rate, and quantization settings optimized for each model size. 

The flow runs on compute nodes equipped with NVIDIA A40-48GB GPUs managed through a SLURM workload scheduler. The compute resources per SLURM job are 16 CPU cores, 160GB system memory, and 2 A40 GPUs. Fine-tuned model checkpoints are loaded from a remote storage system, and data is processed in configurable chunks to maintain memory efficiency and consistent processing speeds. Input data includes standalone Verilog modules and optional C++ code generated by Verilator.

\subsection{Results and Discussion} \label{4d}

\subsubsection{Dataset Labeling Ablation Studies}
\label{subsec:DataAbla}


\begin{table}[H]
    \centering
    \caption{Performance metrics for base LLMs on the VerilogEval-Machine Benchmark}
    \resizebox{0.49\textwidth}{!}{%
    \begin{tabular}{|l|c|c|c|c|}
        \hline
        \textbf{Base LLMs} & \textbf{Release date} & \textbf{Pass@1} & \textbf{Pass@5} & \textbf{Pass@10} \\
        \hline
        Granite3.3-8B & 2025-04-16 & 0.120 & 0.200 & 0.227 \\
        \hline
        Qwen2.5-7B & 2024-09-18 & 0.047 & 0.165 & 0.243 \\
        \hline
        Qwen2.5-32B & 2024-11-11 & 0.526 & 0.642 & 0.670 \\
        \hline
        Claude-3.7 Sonnet & 2025-02-24 & 0.119 & 0.146 & 0.154 \\
        \hline
        GPT-3.5-turbo & 2024-01-25 & 0.076 & 0.187 & 0.238 \\
        \hline
        GPT-4o & 2024-08-06 & 0.503 & 0.599 & 0.621 \\
        \hline
    \end{tabular}%
    }
    \label{tab:non_finetuned_comparison}
\end{table}

\begin{table*}[]
    \centering
    \caption{Comparison of different models and configurations for our fine-tuning ablation study using a downsampled 1k dataset. ``Verilog" means we use all samples in the downsampled 1k dataset that are all re-labeled using our flow without using Verilator; ``Verilog \& C++" means we use all samples in the downsampled 1k dataset relabeled with our flow where 40\% of the modules (all those that pass Verilator) are augmented with C++; ``Verilator Passed" means we only include the 40\% of the downsampled 1k dataset that passed Verilator (and are augmented with C++). Bold numbers represent the best among \doweaccept{``Verilog",} ``Verilog \& C++", and ``Verilator Passed" under the same quantization and language model.}
    \resizebox{0.95\textwidth}{!}{%
    \begin{tabular}{|l|l|c|c|c|c|c|c|c|c|c|}
        \hline
        \textbf{Language Model} &  & \multicolumn{3}{c|}{\textbf{Pass@1}} & \multicolumn{3}{c|}{\textbf{Pass@5}} & \multicolumn{3}{c|}{\textbf{Pass@10}} \\
        \hline
        & & Verilog & Verilog & Verilator & Verilog & Verilog & Verilator & Verilog & Verilog & Verilator \\
        & & & \& C++ & Passed & & \& C++ & Passed & & \& C++ & Passed \\
        \hline
        Granite3.3-8B & INT4 & 0.356 & 0.367 & \textbf{0.370} & \textbf{0.502} & 0.500 & 0.457 & \textbf{0.546} & 0.543 & 0.481 \\
                      & BF16 & \textbf{0.393} & 0.391 & 0.379 & 0.504 & \textbf{0.516} & 0.512 & 0.547 & \textbf{0.565} & 0.540 \\ 
        \hline
        Qwen2.5-7B & INT4 & 0.403 & \textbf{0.499} & 0.426  & 0.532 & \textbf{0.608} & 0.584 &0.576 & \textbf{0.640} & 0.624\\
                    & BF16 & 0.401 & \textbf{0.453} & 0.350 & 0.532 & \textbf{0.614} &0.467 & 0.567 & \textbf{0.655} & 0.524 \\
        \hline
       Qwen2.5-32B & INT4 & \textbf{0.643} & 0.595 &0.566& \textbf{0.746} & 0.728 & 0.692 &\textbf{0.772} & 0.767 &  0.736\\
                    & BF16 & \textbf{0.625} & 0.542 & 0.567 & \textbf{0.723} & 0.672 & 0.674 & \textbf{0.757} & 0.704 & 0.713\\
        \hline
    \end{tabular}%
    }
    \label{tab:performance_comparison}
\end{table*}

\begin{table*}[h!]
    \centering
    \caption{Comparison of language models on VerilogEval using 11k dataset configurations. Datasets: (1) MG-Verilog Original: original Verilog modules with their LLaMA-series generated descriptions; (2) MG-Verilog Re-labeled: identical Verilog modules with descriptions generated by labeling flow; (3) \textsc{OpenRTLSet} Downsampled: modules from our dataset downsampled to 11k samples with our generated descriptions. Labeling configurations include Multi-granularity (detailed, high-level, and block-level descriptions), Single-granularity (detailed descriptions only), and either Verilog-only or Verilog with C++ augmentation.}
    \resizebox{\textwidth}{!}{%
    \begin{tabular}{|l|l|l|c|c|c|c|c|c|}
        \hline
        \textbf{Language Model} & \textbf{Dataset} & \textbf{Labeling} & \multicolumn{3}{c|}{\textbf{VerilogEval-Machine}} & \multicolumn{3}{c|}{\textbf{VerilogEval-Human}} \\
        \hline
        & & & Pass@1 & Pass@5 & Pass@10 & Pass@1 & Pass@5 & Pass@10 \\
        \hline
        Granite3.3-8B & MG-Verilog Original & Multi-granularity & 0.445 & 0.642 & \textbf{0.693} & 0.132 & 0.251 & 0.300 \\
        & & Single-granularity & 0.409 & 0.614 & 0.679 & \textbf{0.206} & 0.314 & 0.357 \\
        & MG-Verilog Re-labeled & Verilog \& C++ & 0.441 & 0.621 & 0.675 & 0.198 & 0.330 & 0.384 \\
        & & Verilog & \textbf{0.470} & \textbf{0.646} & 0.691 & 0.165 & 0.303 & 0.366 \\
        & \textsc{OpenRTLSet} Downsampled & Verilog \& C++ & 0.431 & 0.613 & 0.664 & 0.191 & 0.327 & 0.377 \\
        & and Labeled & Verilog & 0.441 & 0.628 & 0.681 & 0.188 & \textbf{0.331} & \textbf{0.389} \\
        \hline
        Qwen2.5-7B & MG-Verilog Original & Multi-granularity & 0.539 & 0.695 & 0.745 & 0.278 & 0.412 & 0.457 \\
        & & Single-granularity & 0.481 & 0.669 & 0.726 & 0.223 & 0.391 & 0.455 \\
        & MG-Verilog Re-labeled & Verilog \& C++ & 0.537 & \textbf{0.701} & \textbf{0.748} & 0.181 & 0.333 & 0.396 \\
        & & Verilog & 0.530	& 0.679 & 0.723 & 0.267 & 0.395 & 0.443 \\
        & \textsc{OpenRTLSet} Downsampled & Verilog \& C++ & \textbf{0.543} & 0.687 & 0.728 & 0.260 & 0.411 & 0.465 \\
        & and Labeled & Verilog & 0.538 & 0.686 & 0.734 & \textbf{0.280} & \textbf{0.415} & \textbf{0.471} \\
        \hline
        Qwen2.5-32B & MG-Verilog Original & Multi-granularity & 0.594 & 0.765 & 0.814 & 0.327 & 0.457 & 0.500 \\
        & & Single-granularity & 0.495 & 0.690 & 0.741 & 0.231 & 0.401 & 0.463 \\
        & MG-Verilog Re-labeled & Verilog \& C++ & 0.623 & 0.803 & 0.836 & 0.297 & 0.474 & 0.530 \\
        & & Verilog & 0.515 & 0.733 & 0.793 & 0.358	& 0.529 & 0.534 \\
        & \textsc{OpenRTLSet} Downsampled & Verilog \& C++ & \textbf{0.632} & \textbf{0.807} & \textbf{0.839} & \textbf{0.383} & \textbf{0.539} & \textbf{0.581} \\
        & and Labeled & Verilog & 0.618	& 0.775 & 0.817 & 0.369	& 0.531 & 0.578 \\
        \hline
    \end{tabular}%
    }
    \label{tab:performance_comparison_cba}
\end{table*}

Table \ref{tab:non_finetuned_comparison} presents the VerilogEval-Machine Pass@k results for base LLMs, encompassing both open-source models and state-of-the-art closed-source reasoning models such as Claude~\cite{anthropic2024claude} and the GPT family~\cite{openai2024gpt4o}. 
Noteably, the 8B-parameter Granite3.3-8B outperforms both Claude 3.7 and GPT-3.5-turbo, proving that a smaller model can beat larger ones.
Among these models, only GPT-4o and Qwen2.5-32B demonstrate substantial performance on Verilog generation tasks. 
These two models establish our baseline for evaluating the effectiveness of our various fine-tuning strategies and dataset configurations.

Table \ref{tab:performance_comparison} shows the VerilogEval-Machine Pass@k results of various configurations of our labeling flow on MG-Verilog Verilog modules on three LLMs in our ablation study.
``Verilog" means we use all samples in the downsampled 1k dataset that are all re-labeled using our flow without using Verilator; ``Verilog \& C++" means we use all samples in the downsampled 1k dataset relabeled with our flow where 40\% of the modules (all those that pass Verilator) are augmented with C++; ``Verilator Passed" means we only include the 40\% of the downsampled 1k dataset that passed Verilator (and are augmented with C++).

Our results indicate smaller LLMs of 7-8B size generally have weaker performances and thereby, lower Pass@k values compared to Qwen2.5-32B, with Pass@10 of up to 0.772 from Qwen2.5-32B, 0.655 from Qwen2.5-7B, and 0.565 from Granite3.3-8B. Comparing Pass@k results of our 2 Qwen and the Granite3.3-8B LLM in Table \ref{tab:non_finetuned_comparison} and Table \ref{tab:performance_comparison}, we see all these 3 LLMs have clearly higher Pass@k rates after fine-tuning on our downsampled relabeled MG-Verilog, which shows fine-tuning significantly improves LLMs' capability of generating correct Verilog designs.

 ``Verilog \& C++" and ``Verilog" are preferred in our labeling flow compared to ``Verilator Passed" on all three LLMs. This indicates that having a large dataset volume is more important than trying different labeling flow configurations.
Table \ref{tab:performance_comparison} shows different LLMs have different preferences on ``Verilog \& C++" vs ``Verilog": 
Granite3.3-8B and Qwen2.5-7B prefer ``Verilog \& C++", while Qwen2.5-32B achieves better results on ``Verilog". 


Our results suggest that INT4 and BF16 have their respective strengths, but ultimately, INT4 is selected to label our dataset using DeepSeek-R1 70B inference to save computational resources. The model Qwen2.5-32B with quantization scheme INT4 and ``Verilog" labeling flow setting has the highest mean Pass@10 of 0.772 as highlighted in the table.

\subsubsection{Comparative Benchmark Analysis}

The comparative benchmark analysis in Section \ref{4b} isolates and quantifies the contributions of the dataset composition and labeling flow to the model performance. 
Table \ref{tab:performance_comparison_cba} shows comparisons of six datasets each with 11k Verilog modules in fine-tuning three LLMs then running VerilogEval: two datasets are from original MG-Verilog, one with just the detailed descriptions as labels (Single-granularity) and the other with all the detailed, high-level, and block-level descriptions as labels (Multi-granularity); two datasets are from relabeled MG-Verilog using our DeepSeek-R1 70B flow, one using ``Verilog \& C++", and the other using ``Verilog", both defined in the same way as the subsection~\ref{subsec:DataAbla}; two datasets are from downsampled \textsc{OpenRTLSet}, also with one using ``Verilog \& C++", and the other using ``Verilog".

We consider the original MG-Verilog dataset~\cite{zhang2024mgverilog} as a baseline with its original LLaMA-series-generated labels. 
As shown in Table \ref{tab:performance_comparison_cba}, the original MG-Verilog dataset in both multi-granularity and single-granularity module description labels mostly yield the worst performance across all Pass@k metrics on all three LLMs.

To assess the impact of our labeling flow, we re-labeled MG-Verilog using our DeepSeek-R1 70B prompting framework. We provide two configurations of our flow: ``Verilog \& C++" and ``Verilog", defined in the same way as the subsection~\ref{subsec:DataAbla}. We use both two configurations to re-label all 11k Verilog modules from MG-Verilog using our flow.
The re-labeled MG-Verilog mostly yields performance gains across all Pass@k evaluation metrics, with the most substantial improvements observed in Pass@10. 

To further analyze the effect of dataset quality, we evaluate a downsampled version of \textsc{OpenRTLSet}, consisting of 11k samples to match the scale of MG-Verilog. This version is also labeled using DeepSeek-R1 70B prompting framework, ensuring a controlled comparison where only the dataset content varies. 
Again, both ``Verilog \& C++" and ``Verilog" are provided.
Our downsampled \textsc{OpenRTLSet} surpasses the original MG-Verilog dataset in all cases. 

\subsubsection{Key Insights and Findings}

Several key insights emerge from our study. 
First, among all the three LLMs, dataset quality has the biggest impact at lower number of trials on VerilogEval, with the most significant improvements observed in Pass@1, increasing by up to 0.164 (e.g., Qwen2.5-32B improving from 0.495 to 0.659). This highlights the importance of well-crafted datasets for generating high-confidence outputs in the first trial.
Second, while re-labeling MG-Verilog significantly narrows the performance gap with \textsc{OpenRTLSet}, the latter still retains a clear edge, confirming that both dataset and label quality contribute to model performance.

\subsection{Results on the Full 131k \textsc{OpenRTLSet}}

In addition to the experiments in Section \ref{4d}, we also performed extensive evaluations of the entire 131k \textsc{OpenRTLSet} on the VerilogEval Benchmark. Table \ref{tab:openrtlset_131k_vs_11k_machine} shows comparisons of the best results in various code enhancement strategies on the downsampled 11k-size \textsc{OpenRTLSet} in Section \ref{4d} vs. the result of the complete 131k \textsc{OpenRTLSet} on the VerilogEval-Machine Benchmark; Table \ref{tab:openrtlset_131k_vs_11k_human} presents the same comparisons as Table \ref{tab:openrtlset_131k_vs_11k_machine} on the VerilogEval-Human Benchmark. As in Sections \ref{4c} and \ref{4d}, we fine-tune two families of models—Granite-Code-Instruct (8B) and Qwen2.5-Coder (7B and 32B). 

\begin{table}[htbp]
    \centering
    \caption{Performance metrics for LLMs finetuned on 11k vs. 131k \textsc{OpenRTLSet} on the VerilogEval-Machine Benchmark}
    \resizebox{0.49\textwidth}{!}{%
    \begin{tabular}{|l|c|c|c|c|}
        \hline
        \textbf{LLMs} & \textbf{Dataset \& Strategy} & \textbf{Pass@1} & \textbf{Pass@5} & \textbf{Pass@10} \\
        \hline
        Granite3.3-8B & Best of 11k Datasets & 0.470 & 0.646 & 0.693 \\
                   & \textsc{OpenRTLSet} 131k    & \textbf{0.471} & \textbf{0.660} & \textbf{0.727} \\
        \hline
        Qwen2.5-7B & Best of 11k Datasets & 0.543 & 0.701 & 0.748 \\           
                   & \textsc{OpenRTLSet} 131k    & \textbf{0.579} & \textbf{0.749} & \textbf{0.802} \\
        \hline
        Qwen2.5-32B & Best of 11k Datasets & 0.632 & 0.807 & 0.839 \\
                    & \textsc{OpenRTLSet} 131k    & \textbf{0.651} & \textbf{0.848} & \textbf{0.893} \\
        \hline
    \end{tabular}%
    }
    \vspace{-2mm}
    \label{tab:openrtlset_131k_vs_11k_machine}
\end{table}

Table \ref{tab:openrtlset_131k_vs_11k_machine} shows significant and consistent improvements in VerilogEval-Machine Pass@k of all three fine-tuned LLMs on the complete 131k \textsc{OpenRTLSet} compared against downsampled 11k \textsc{OpenRTLSet}, with up to 3.6\% absolute gains in Pass@1, up to 4.8\% absolute gains in Pass@5, and up to 5.4\% absolute gains in Pass@10. All three fine-tuned LLMs have $>$3\% absolute gains in Pass@10. 
\doweaccept{Among all three evaluated LLMs, Qwen2.5-7B achieves the largest gains among the three VerilogEval-Machine Pass@k metrics when comparing fine-tuning on the complete 131k OpenRTLSet with fine-tuning on the best 11k dataset.}
Remarkably, Qwen 2.5-32B LLM achieves 89.3\% of Pass@10 without using any additional inference-time enhancements like RAG or CoT. 

\begin{table}[htbp]
    \centering
    \caption{Performance metrics for LLMs finetuned on 11k vs. 131k \textsc{OpenRTLSet} on the VerilogEval-Human Benchmark}
    \resizebox{0.49\textwidth}{!}{%
    \begin{tabular}{|l|c|c|c|c|}
        \hline
        \textbf{LLMs} & \textbf{Dataset \& Strategy} & \textbf{Pass@1} & \textbf{Pass@5} & \textbf{Pass@10} \\
        \hline
        Granite3.3-8B & Best of 11k Datasets & 0.206 & 0.331 & 0.389 \\
                   & \textsc{OpenRTLSet} 131k    & \textbf{0.207} & \textbf{0.344} & \textbf{0.391} \\
        \hline
        Qwen2.5-7B & Best of 11k Datasets & 0.280 & 0.415 & 0.471 \\
                   & \textsc{OpenRTLSet} 131k    & \textbf{0.284} & \textbf{0.456} & \textbf{0.519} \\
        \hline
        Qwen2.5-32B & Best of 11k Datasets & 0.383 & 0.539 & 0.581 \\
                    & \textsc{OpenRTLSet} 131k    & \textbf{0.424} & \textbf{0.587} & \textbf{0.642} \\
        \hline
    \end{tabular}%
    }
    \vspace{-2mm}
    \label{tab:openrtlset_131k_vs_11k_human}
\end{table}

Table \ref{tab:openrtlset_131k_vs_11k_human} also shows consistent improvements in VerilogEval-Human Pass@k in all three fine-tuned LLMs on the \textsc{OpenRTLSet} 131k compared against downsampled \textsc{OpenRTLSet} 11k, with up to 4.1\% absolute gains in Pass@1 ratio, up to 4.8\% absolute gains in Pass@5 ratio, and up to 6.1\% absolute gains in Pass@10 ratio. 
\doweaccept{Among all three evaluated LLMs, Qwen2.5-32B achieves the largest gains across all VerilogEval-Human Pass@k metrics when comparing fine-tuning on the complete 131k OpenRTLSet with fine-tuning on the best 11k dataset.}

\section{Conclusion}
\label{section:conclusion}
There is significant interest in leveraging LLMs to aid in the development of hardware designs. However, there are relatively fewer open-source hardware designs compared to software. Furthermore, LLMs still face significant challenges in generating correct hardware designs. 
\textsc{OpenRTLSet} is the largest fully open-source Verilog dataset comprised of more than 131,000 unique Verilog modules each with high-quality natural language description. Each Verilog module is labeled and linked to the original source repository. We conducted extensive studies on our flow and dataset to evaluate their effectiveness and provide insight into possible future research directions. 
Our \textsc{OpenRTLSet} will help pave the way towards more comprehensive and powerful LLM tools for hardware design in both academic and commercial use cases.

\section{Acknowledgments}
\doweaccept{
This work is supported by the AMD Center of Excellence, the AMD HACC Initiative, the SRC grant 2023-CT-3175, and the NSF ACCESS computing resources. 
}

\doweaccept{
This work leverages the National Center for Supercomputing Applications (NCSA) Delta cluster. We are supported by the National Science Foundation (award OAC 2005572) and the State of Illinois.
}

\bibliography{reference}

\end{document}